\documentclass[manuscript,screen]{acmart}
\usepackage{algorithm,algpseudocode}
\usepackage{enumitem}
\AtBeginDocument{%
  \providecommand\BibTeX{{%
    \normalfont B\kern-0.5em{\scshape i\kern-0.25em b}\kern-0.8em\TeX}}}

\begin{document}
\title{Complete Implementation of WXF Chinese Chess Rules}


\author{Daniel Tan}
\affiliation{%
  \institution{University of California, Riverside}
  \city{Riverside}
  \country{United States of America}}
\email{dtan004@ucr.edu}

\author{Neftali Watkinson Medina}
\affiliation{%
  \institution{University of California, Riverside}
  \city{Riverside}
  \country{United States of America}
}
\email{neftaliw@ucr.edu}

\renewcommand{\shortauthors}{Daniel Tan and Neftali Watkinson Medina}

\begin{abstract}
Unlike repetitions in Western Chess where all repetitions are draws, repetitions in Chinese Chess could result in a win, draw, or loss depending on the kind of repetition being made by both players. One of the biggest hurdles facing Chinese Chess application development is a proper system for judging games correctly. This paper introduces a complete algorithm for ruling the WXF rules correctly in all 110 example cases found in the WXF manual. We introduce several novel optimizations for speeding up the repetition handling without compromising the program correctness. This algorithm is usable in engines, and we saw a total increase in playing strength by +10 point rating increase, or an increased 5\% winrate when integrating this approach into our prototype engine.
\end{abstract}




\maketitle

\section{Introduction}

Chinese Chess (or Xiangqi) is one of the most popular board games in Asian countries such as China and Vietnam [15]. Chinese Chess is an indispensable part of the daily lives of Chinese people. There are over one hundred Chinese idioms that have their roots in Chinese Chess [15]. Chinese Chess and Western Chess share many similarities. Both games are two player turn-based strategy games where the objective of the game is to checkmate the opponent's king using a set of pieces with varying kinds of movement. However, while in Western Chess having multiple repetitions can end in a draw, in Chinese Chess repetitions can result in a win, loss, or draw depending on how that repetition is performed. Judging the actual outcome of repetitions is a complex, nontrivial problem.  In this paper, we implement and document an algorithm that completely covers all 110 example diagrams. Unlike other implementations that are missing several cases [12], we present an algorithm that correctly identifies each case correctly.

\subsection{Overview of Chinese Chess Repetition Rules}

\noindent A short overview of the repetition rules is that no player is allowed to either perpetually chase an unprotected opponent's piece or perpetually check the opponent's king. The player that performs a perpetual chase or a perpetual check loses the game. This is completely different than in Western Chess where a perpetual check is a draw and perpetual chases are close to nonexistent. Under all conditions, perpetual check in Chinese Chess results in a loss for the player constantly checking the opponent's king. If both players simultaneously perform perpetual checks, then the game will result in a draw. If both players simultaneously perform perpetual chases, then the game will result in a draw. If neither player is doing a perpetual check nor a perpetual chase, then the game is ruled a draw. Figures 1, 2, and 3 provide an example of perpetual check, perpetual chase, and both players simultaneously performing perpetual check, respectively.\\*

\begin{figure}[!tbp]
  \centering
  \begin{minipage}[b]{0.3\textwidth}
    \includegraphics[width=\textwidth]{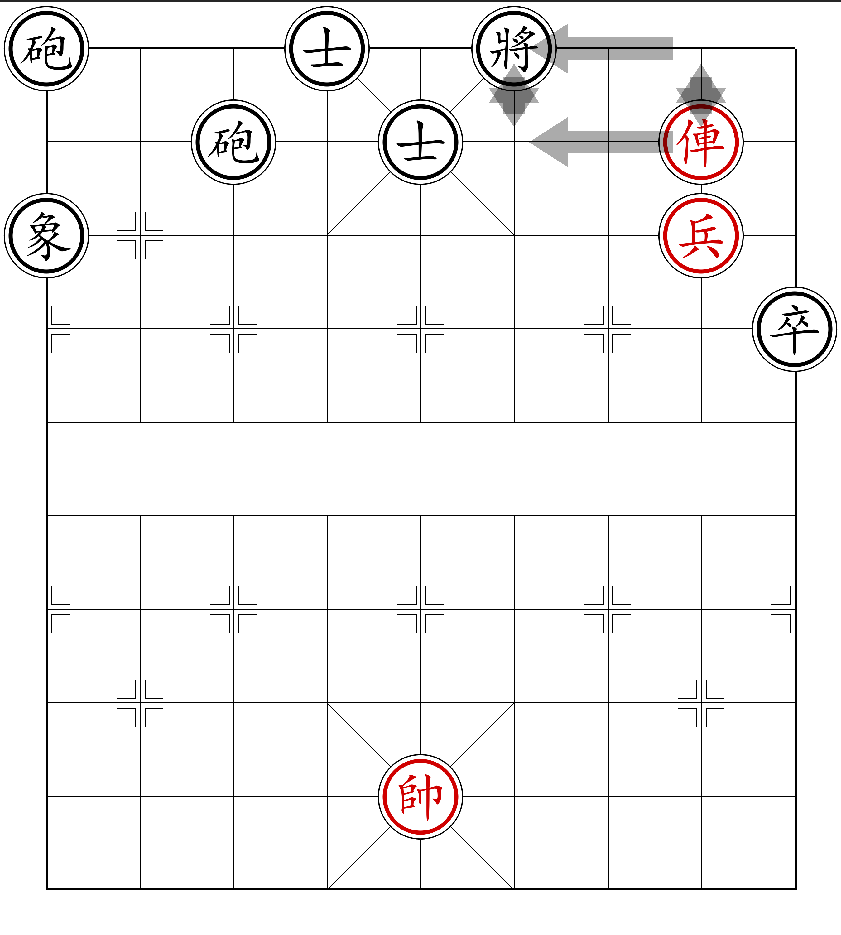}
    \caption{The red rook is perpetually checking the black king R2+1, K6+1, R2-1, K6-1, R2+1, K6+1, R2-1...red loses if he continues to check the king.}
  \end{minipage}
  \hfill
  \begin{minipage}[b]{0.3\textwidth}
    \includegraphics[width=\textwidth]{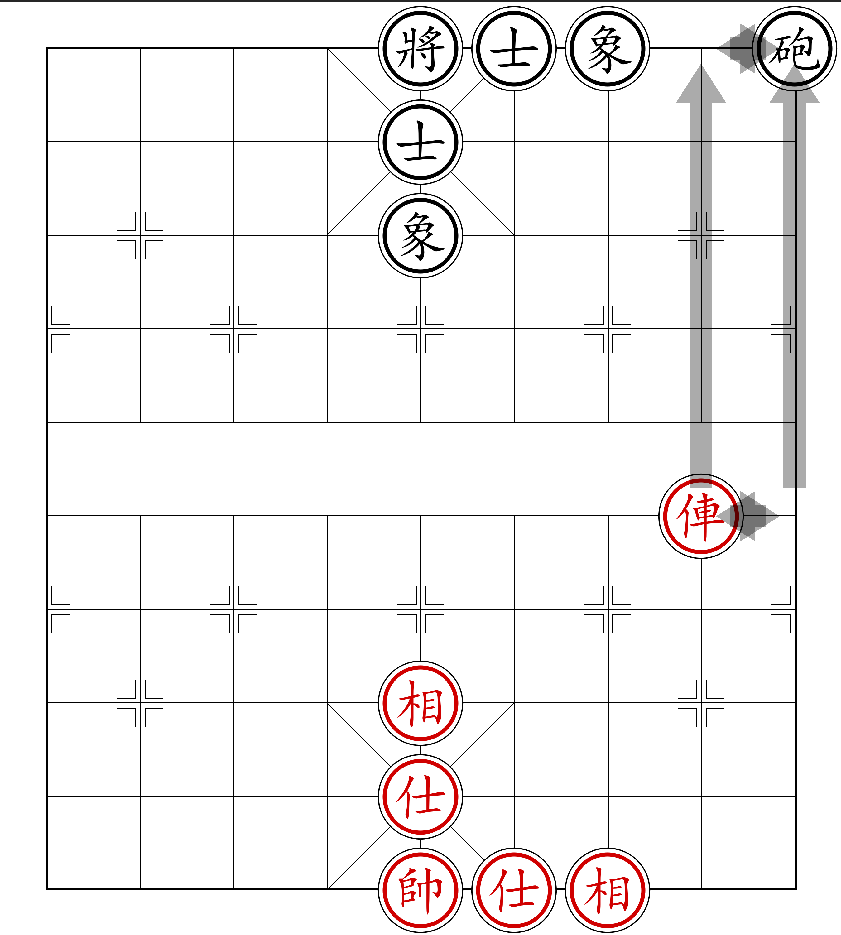}
    \caption{The red rook is perpetually chasing the unprotected black cannon R2=1, C9=8, R1=2, C8=9 ...red loses if he continues to chase the cannon.}
  \end{minipage}
  \hfill
  \begin{minipage}[b]{0.3\textwidth}
    \includegraphics[width=\textwidth]{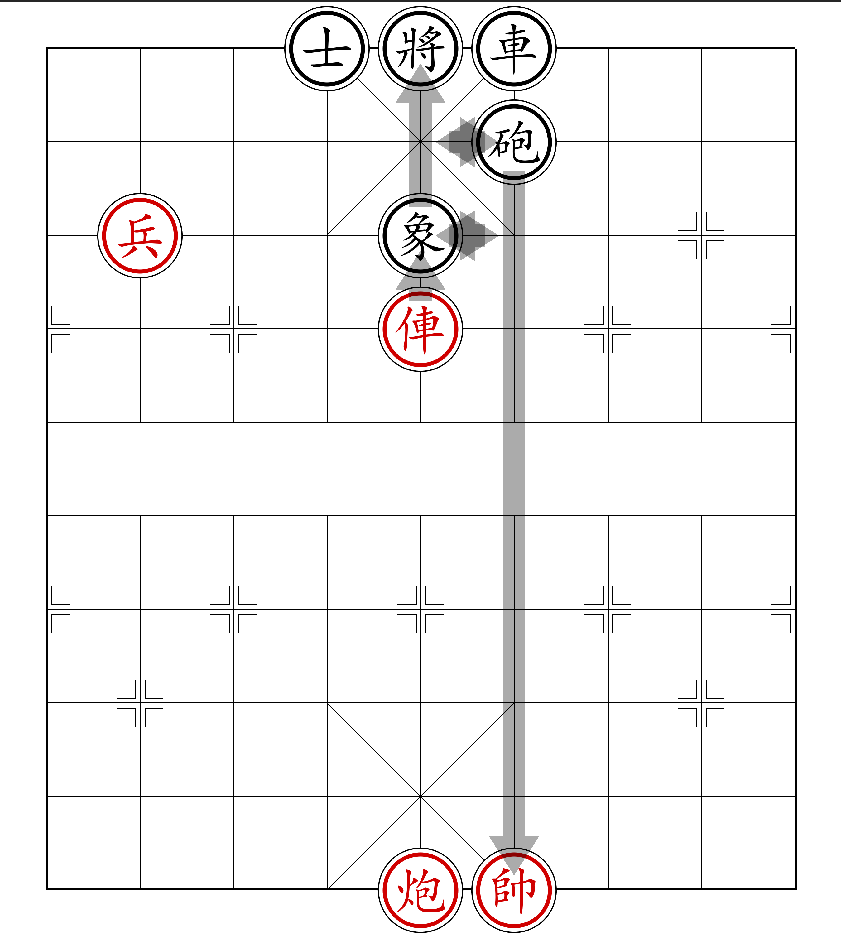}
    \caption{Both players perpetually check each other simultaneously R5+1, C5=6, R5=4, C6=5, R4=5, C5=6...this repetition leads to a draw}
  \end{minipage}
\end{figure}

\noindent Most Chinese Chess players, from casual enthusiasts to top grandmasters, have an intuitive grasp of the repetition rules, but do not fully understand all the repetition rules [12]. The whole set of rules to determine who wins, draws or loses is complex. This is even more complex if both players are engaging in a chase or if multiple pieces are chasing multiple pieces. If a tournament game runs into a dispute, a special qualified referee is called to resolve the dispute. If a referee determines that a player is performing a losing repetition, the referee will warn the losing side to change moves or lose the match.\\*

\noindent There are two main versions of the official Chinese Chess repetition rules used in tournament play: the WXF rules (World Xiangqi Federation Rules) [21] and the CXA rules (Chinese Xiangqi Association Rules) [3]. The WXF rules are used for international competitions between different countries, while the CXA rules are used in Mainland China. The CXA rules are complicated, subjective, and contradictory [23], but games using these rules are less likely to end with a draw. The WXF rules lead to more draws, but are more straightforward and consistent. This paper focuses on WXF rules because CXA rules are too subjective and WXF is the standard for Chinese Chess engine play and international competitions between different countries.\\*

\noindent The official WXF rules [21] are summarized in 4 pages (WXF manual 35-39), with around 110 example diagrams with paragraphs of explanations for each diagram spanning 59 pages (WXF manual 40-99). Given that the WXF manual is 116 pages long, over two thirds of the manual is dedicated to outcomes for repetitions. In this paper, we implement and document an algorithm that completely covers all 110 example diagrams. Unlike other implementations that are missing several cases [12], we present an algorithm that correctly identifies each case correctly. When we integrated this algorithm into our game engine, the algorithm slightly improved the engine playing strength without any significant slowdown.

\subsection{Motivation}

Due to the sheer complexity of the rules, Chinese Chess programmers put more emphasis on evaluating millions of nodes as fast as possible rather than rule correctness [12]. Heuristics are used to estimate repetitive chasing accurately for a majority of judgement calls. Errors in heuristics can be compensated by having a good minimax search and/or better handcrafted evaluation heuristics.\\*

\noindent Despite this, a complete implementation is still important. For example, Chinese Chess websites and software would benefit from a complete implementation of the rules since it is impossible to get a human referee to oversee all possible online or virtual games. People playing online should not be able to swindle a victory or a draw by exploiting losing repetition moves that the software is not able to detect.  If even a few example cases are wrong, it could lead to many thousands of games ruled incorrectly and even threaten the program's stability.\\*

\noindent To illustrate this point, consider diagram 34 (figure 4) from the WXF manual [21]. Black is attempting to checkmate red by moving the cannon to the bottom rank, while red attempts to block the black cannon using the red rook. Black moves back and forth, perpetually threatening checkmate, while red perpetually threatens the unprotected black cannon. The black perpetual kill is allowed, while the red perpetual chase results in a loss for red. If red breaks the repetition, he will end up getting checkmated by the black cannon. But if red continues to perform the repetition, he will lose due to the perpetual chase rule. Either way, red has lost. These situations are not just hypothetical; grandmaster strategy often involves forcing opponents into critical situations similar to this to either force a significant loss of material or deliver a checkmate [22]. Different software implementations have varying degrees of support for the rules. Many popular Chinese Chess software applications [14] fail to support cases such as this, but others such as [6], [12], and [13] handle this case correctly.\\*

\begin{figure}[!tbp]
  \centering
    \includegraphics[width=0.3\textwidth]{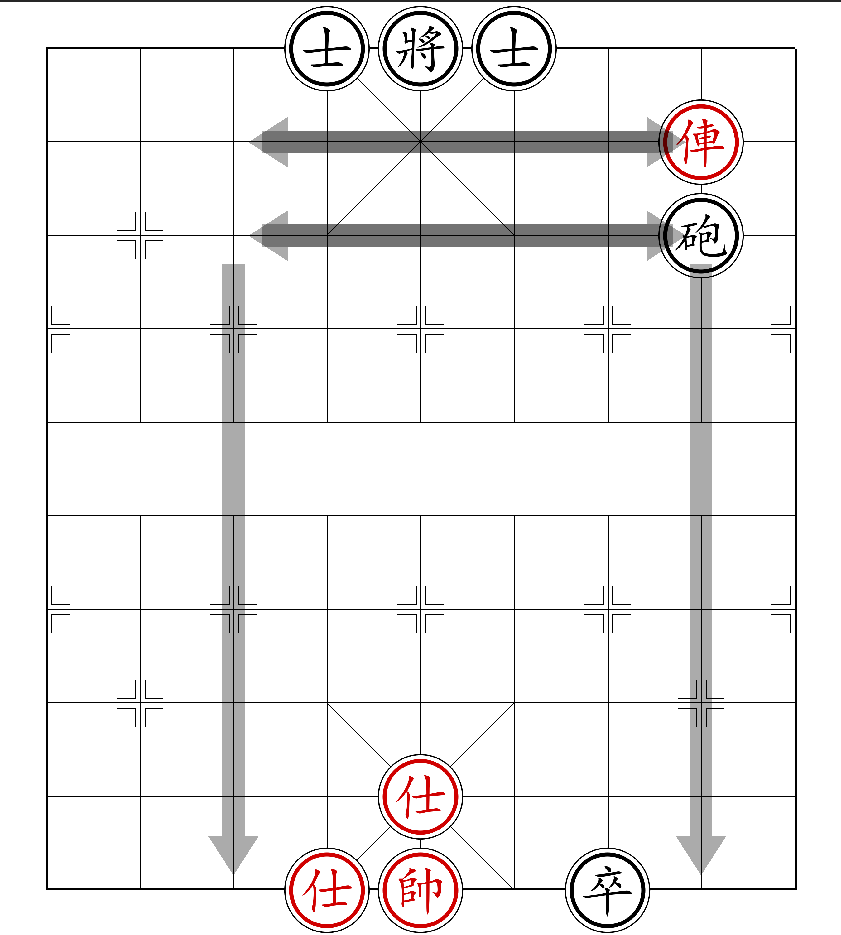}
    \caption{Both players make the repetition c8=3, R2=7, c3=8, R7=2... The black cannon repeatedly attempts to perpetually kill (attempting to checkmate) the red king. The red rook perpetually chases the black cannon. Red will lose by repetition if he continues repeating moves.}
\end{figure}

\noindent Therefore, one of the biggest hurdles facing Chinese Chess application development is a proper system for judging games correctly. In the most popular Chinese Chess application Tiantian Xiangqi, repetitions are not handled in the software at all, and most of the time, it rules repetitions incorrectly as draws [14]. Most repetition handling in the most popular Chinese Chess applications are completely broken.\\*

\noindent Another motivating factor is that with the introduction of techniques such as NNUE: Efficiently updatable neural networks [10], handcrafted evaluation functions are being replaced with more dynamic neural network evaluation functions that play in a more human like way. NNUE, unlike handcrafted evaluation, has millions of parameters all tuned automatically using stochastic gradient descent. A handcrafted evaluation may have only several hundred parameters that can be easily hand-tuned by a programmer, but a neural network with millions of parameters is impossible to tune by hand. From our observations during NNUE training, the model exaggerated the errors created by the heuristics, led to poor decision making, and resulted in a decrease in playing strength.\\*

\noindent And finally, a complete implementation is necessary for Alpha-Zero [16] like engines using Monte Carlo Tree Search, where an engine learns the game from scratch without any human intervention. When one is building a self-learning Chinese Chess engine starting from zero knowledge, it is necessary to be able to evaluate all positions accurately to get the correct win/draw/loss ratio for Monte Carlo Tree Search to function correctly.

\section{Background}

Very little work has been done regarding a full implementation of Chinese Chess rules. Most engines have partial solutions designed for speed rather than correctness. Even implementations advertised to cover all repetition rules are off by 3-5 example cases, leading some to argue that support for all WXF rules is an impossible endeavor [12]. In this paper, we demonstrate that all WXF rules can be supported, and that getting all diagrams perfectly correct is possible. Getting all diagrams correct in all cases is important: if one example case is wrong, it could lead to many thousands of games ruled incorrectly.\\*

\noindent Given that the vast majority of repetitions are perpetual checks [12], the simplest and most effective solution to the Chinese Chess repetition rules is to punish the player that is doing the perpetual check with a loss, while all other repetitions are marked as draws. Under this scheme, all losing perpetual chases get incorrectly labeled as draws. Simple open source didactic engines such as [25] handle repetitions in this way. At the average Chinese Chess player level and basic engines, perpetual chases do not affect gameplay as much. Perpetual chase handling only start impacting strategies at the grandmaster level.\\*

\noindent An improvement upon just perpetual check handling is one described in [23]. In their Chinese Chess engine Contemplation, they do perpetual check handling, then if a cycle matches a frequently encountered chasing pattern, they report the correct score that complies with the Chinese Chess rules. The algorithm returns return a low value of -700 when it encounters an unhandled case, punishing the player making the repetition. This is correct enough in most cases when repeating moves is bad. However, returning -700 causes the engine to fail to understand drawing or winning by repetition and cause the program to act abnormally. They do not provide any data regarding whether their improved engine has higher playing strength compared to their previous version.\\*

\noindent Xiexiemaster [18] is a strong grandmaster level engine created by programmer Pascal Tang. Xiexiemaster won many Computer Chinese Chess tournaments, including getting 1st place in the Computer Chinese Chess World Championship in 2004. Xiexiemaster earned fame within the Chinese Chess community after beating several top grandmasters and international masters. According to their documentation, Xiexiemaster software understands 95\% of the complex Chinese Chess repetition rules.\\*

\noindent Pikafish [13] is an open source Chinese Chess engine forked from state of the art Western Chess Engine Stockfish [17]. Pikafish took the Stockfish search algorithm and NNUE evaluation code for Western Chess and modified the code to handle Chinese Chess. The developers focus on handling all repetition cases correctly at the cost of slowing down the engine. Most cases are handled correctly most of the time. However, Pikafish version 2023-12-03 fails diagrams 39, 40, 41 as well as the final Mutual Perpetual Chase diagram on the final page. They cover 106 out of 110 cases found in the WXF manual.\\*

\noindent [12] attempts to implement all Chinese Chess rules and integrated their approach into their Felicity Chinese Chess engine. They have 106 examples working correctly, but fails the same 4 cases Pikafish fails, diagrams 39, 40, 41, and 61. After implementing a solution where a cannon or rook is not considered making a clear protection on a piece, their solution fixed almost all cases, except for diagram 108, leaving the author with 109 out of 110 games correct. Their solution is more accurate than Pikafish, but their solution remains incomplete.\\*

\noindent Unlike previous work, we present a solution that handles all WXF example cases with 100\% accuracy. No missing cases, and a unique algorithm for correctly handling all diagrams; all 110 out of 110 cases work correctly. In the implementation details section, we will go in-depth about why their solutions failed to detect the repetition correctly while ours did. Our implementation increased the playing strength of our engine without compromising on correctness.

\section{Definitions}
Due to the technical nature of this paper, we want to establish some clear terminology and how we define words. Since we implement WXF rules [22], we borrow heavily from their terminology. For the sake of clarity, we provide the following definitions:\\*

\noindent \textbf{attacker} - The piece that is performing an attack on another piece. If a rook attacks an unprotected cannon, for example, the rook is the attacker.\\*

\noindent \textbf{capture move} - a move that captures a piece\\*

\noindent \textbf{check} - a condition that occurs when a player's king is under attack\\*

\noindent \textbf{chase} - a move that threatens to capture a piece other than the king in the next move.\\*

\noindent \textbf{exchange} - a move that trades a piece with an opponent's piece (e.g. trading rook for rook).\\*

\noindent \textbf{idle} - a move that is not a check or chase. We are using the term 'idle' slightly different than the WXF manual [22] terminology. While the WXF manual makes a distinction between terminologies such as 'kill', 'block', 'offer', 'idle', etc., our algorithm makes no such distinction. In software, all these different terms are lumped as an 'idle'.\\*

\noindent \textbf{is\_legal} - this function determines whether a piece or king can legally move to a square without exposing the king to capture by an opponent's piece.\\*

\noindent \textbf{is\_pseudolegal} - this function determines whether a piece can move to a square regardless of whether the move would put the king in check. A valid pseudo-legal move could be illegal due to a pin on the king.\\*

\noindent \textbf{quiet move} - a move that does not capture a piece\\*

\noindent \textbf{river} - refers to the portion of the board between the fifth and sixth rank. The river divides the board into the two halves: one side for each player.\\*

\noindent \textbf{victim} - The piece being attacked. If a rook attacks an unprotected cannon, the cannon is the victim.\\*

\section{Implementation Details}

In this portion of the paper, we will explain in detail how the repetition rules work, how we interpret and frame the problem, and how that framing influences the design of our algorithm.\\*

\noindent There are four main parts of the algorithm: generating captures on unprotected pieces, evaluating whether a particular capture is a chase, determining the violation level of a single player, and a function that returns whether the position has no repetitions, is a winning/losing repetition relative to the side, or a draw. There are a few basic fundamental functions and data structures we will also explain in detail.

\subsection{Data Structures Problem Representation}

In our Chinese Chess position data structure, we create a position history data structure for storing position hashes of previous positions, the previous moves, a position status enum that could be either a chase, check, or idle, and a set of chased pieces. We compute position hashes using techniques described in [26]. When determining if the position is repeating, we traverse backward from the current position examining previous position hashes, and if the hash value of the position matches the hash value of a previous board position found in the position history data structure, then it is considered a repetition. If there is a chase, we track the set of consistently chased pieces to detect whether a perpetual chase is happening or not. The algorithm uses a bitwise \textbf{or} on all previous board position status, and enums (enumerators) to determine whether all positions are checking, chasing, idling, or some hybrid of both.

\begin{algorithm}
\begin{algorithmic}[1]
\caption{Data Structures}\label{Data Structures}

\algblock[Name]{Struct}{EndStruct}
\algblockdefx[NAME]{START}{END}%
[2][Unknown]{Start #1(#2)}%
{Ending}
\algblockdefx[NAME]{}{OTHEREND}%
[1]{Until (#1)}

\algblock[Name]{Enum}{EndEnum}
\algblockdefx[NAME]{START}{END}%
[2][Unknown]{Start #1(#2)}%
{Ending}
\algblockdefx[NAME]{}{OTHEREND}%
[1]{Until (#1)}

\Struct{Info}
  \State integer32 hash,
  \State Move move,
  \State Status status,
  \State Chased\_Pieces chased\_set = $\emptyset$,
\EndStruct
\State
\Enum{Status}
\State Idle = 1, Chase = 2, Check = 4
\EndEnum
\State
\Struct Position
\State ...all other position data
\State Info [] position\_history,
\EndStruct

\end{algorithmic}
\end{algorithm}

\subsection{Protected Pieces}

The \textit{is\_protected} function determines whether a piece is attacking an unprotected piece. The implementation of \textit{is\_protected} matches the definition found in the WXF manual [22], which is if at least one defender can recapture the piece in the next move legally, then return true. If no defenders can come to the defense of the piece, then return false. If a less valuable piece is attacking a more valuable piece (e.g. cannon attacks rook, knight attacks rook), then \textit{is\_protected} returns false, since it is almost always advantageous to exchange a less valuable piece for a more valuable piece. This function could be considered a simplified version of a static exchange evaluation function [9], except instead of returning an evaluation numerical value based on whether a series of exchanges is beneficial or not, it returns a boolean value based on whether a piece is protected.\\*

\begin{algorithm}
\begin{algorithmic}
\caption{is\_protected}\label{is_protected}
\Procedure{is\_protected}{$position$, $move$} \Comment{Determines whether a victim is protected}
\If {$piece\_value\_of(move.to) > piece\_value\_of(move.from)$}
      \State \Return $False$ \Comment{return false if victim is greater than attacker}
\EndIf
\State $occupied \gets position.occupied\_pieces$ \Comment{Get the occupied pieces data of the current position}
\State \Call{make\_move}{position, move}
\ForAll {$move \in $ \Call{get\_defenders}{position, $move.to$, $occupied$}} \Comment{Loop through all possible defenders}
  \If {\Call{is\_legal}{position, $move.to$}}
    \State \Call{unmake\_move}{position, move}
    \State \Return $True$
  \EndIf
\EndFor
\State \Call{unmake\_move}{position, move}
\State \Return $False$
\EndProcedure
\end{algorithmic}
\end{algorithm}

\begin{algorithm}
\begin{algorithmic}
\caption{get\_defenders}\label{is_protected}
\Procedure{get\_defenders}{$position$, $to$, $occupied$} \Comment{Find pieces that can defend the square "$to$"}

\If {\Call{get\_advisor\_move}{$to$}}
      \State \Return $advisor\_move$
\EndIf

\If {\Call{get\_elephant\_move}{$to$}}
      \State \Return $elephant\_move$
\EndIf

\If {\Call{get\_pawn\_move}{$to$}}
      \State \Return $pawn\_move$
\EndIf
\If {\Call{get\_knight\_move}{$to$}}
      \State \Return $knight\_move$
\EndIf

\If {\Call{get\_cannon\_move}{$to$, $occupied$}} \Comment{Determine cannon capture moves based off 'occupied' variable}
      \State \Return $cannon\_move$
\EndIf

\If {\Call{get\_rook\_move}{$to$, $occupied$}} \Comment{Determine rook capture moves based off 'occupied' variable}
      \State \Return $rook\_move$
\EndIf

\If {\Call{get\_king\_move}{$to$}}
      \State \Return $king\_move$
\EndIf
\EndProcedure
\end{algorithmic}
\end{algorithm}

\noindent The formal WXF definition of a protected piece can be tricky, and an incorrect definition of protection in [12] and [13] causes these engines to fail to support all the WXF rules. Protection is determined from the current position rather than subsequent moves that may change the protection status of a piece. In figures 5 and 6, because the black advisor and black chariot are in between the red knight and cannon, the red cannon is considered protecting the red knight. The subsequent advisor move is not considered removing the protection of the knight. Black is alternating between blocking the cannon attack and another turn chasing the red knight with the pinned rook. Red is perpetually chasing the black chariot on every move. Red loses since red is doing a perpetual chase while black is alternating between blocking and chasing. (figures 5,6, and 7).\\*

\noindent The \textit{get\_defenders} subfunction used within the \textit{is\_protected} function iterates through all possible defending pieces, and if a defender is found, returns a move that defends the piece with a recapture. The cannon and rook defending a piece is determined from the position and pieces occupied in the current position in consideration rather than the subsequent move that affects how protection works.\\*

\begin{figure}[!tbp]
  \centering
  \begin{minipage}[b]{0.3\textwidth}
    \includegraphics[width=\textwidth]{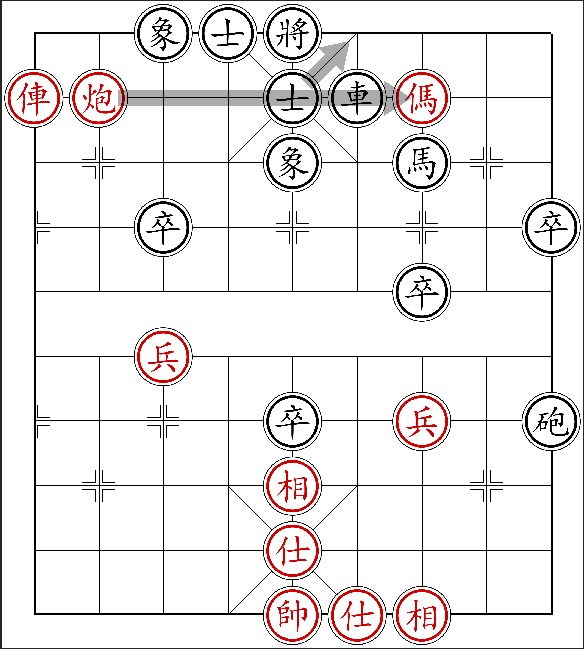}
    \caption{Diagram 40 from WXF manual: The red cannon attacks the black chariot and protects the red knight. The black chariot is not considered attacking the red knight, and moving the black advisor to make the red knight undefended is not considered a chase.}
  \end{minipage}
  \hfill
  \begin{minipage}[b]{0.3\textwidth}
    \includegraphics[width=\textwidth]{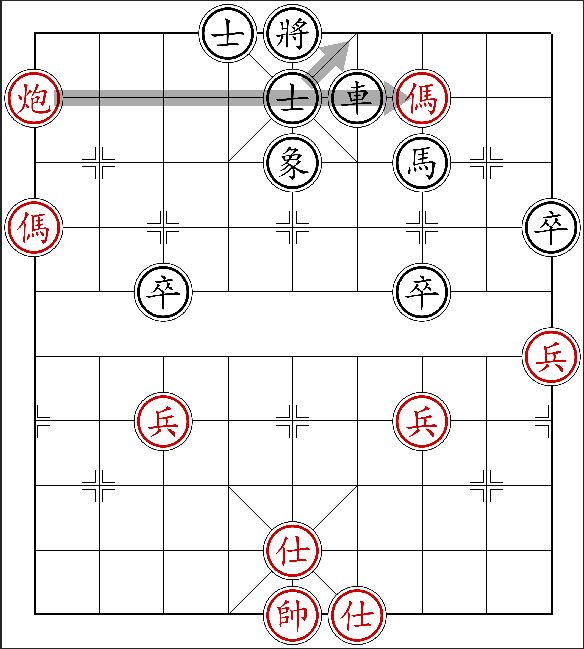}
    \caption{Diagram 41 from the WXF manual: The red cannon attacks the black chariot while protecting the red knight. Similar to diagram 40, moving the black advisor to make the red knight undefended is not considered a chase and black is not considered chasing.}
  \end{minipage}
  \hfill
  \begin{minipage}[b]{0.3\textwidth}
    \includegraphics[width=\textwidth]{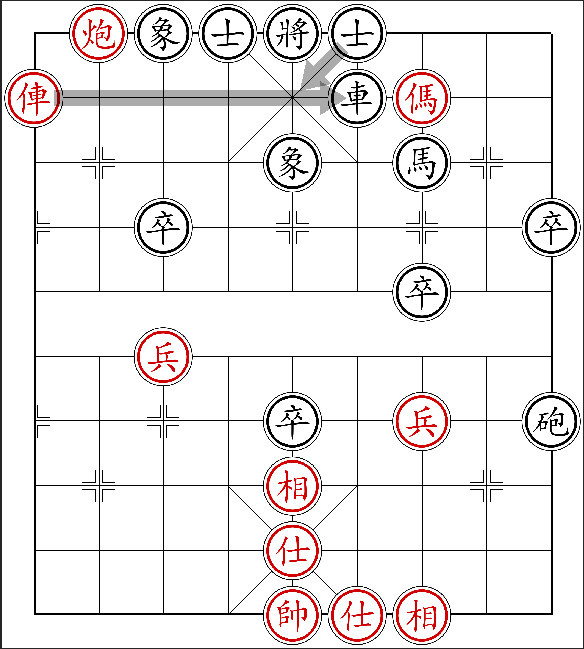}
    \caption{Diagram 41 from the WXF manual: The red chariot attacks the black chariot. If the black advisor moves to cut off the red chariot attack, the black chariot is performing an attack on the red knight. This case should be considered a chase by the black chariot.}
  \end{minipage}
\end{figure}

\subsection{Generating Captures on Unprotected Pieces}

\noindent When generating captures on unprotected pieces, we generate all pseudo-legal capture moves of the opposing side. We then filter out all illegal moves and captures on protected pieces.\\*

\begin{algorithm}
\begin{algorithmic}
\caption{gen\_captures}\label{gen_captures}
\Procedure{gen\_captures}{$position$} 
\State $moves \gets $ \Call{create\_move\_list}{}
\Comment{Skip king moves and pawn moves}
\State \Call{generate\_knight\_moves}{$position$, $moves$}
\State \Call{generate\_rook\_moves}{$position$, $moves$}
\State \Call{generate\_cannon\_moves}{$position$, $moves$}
\State \Call{generate\_advisor\_moves}{$position$, $moves$}
\State \Call{generate\_elephant\_moves}{$position$, $moves$}
\State \Return $moves$
\EndProcedure
\end{algorithmic}
\end{algorithm}

\noindent There are many exceptions to attacks on unprotected pieces. If a lone king or pawn chases a piece, the game will be ruled as a draw. Exchange (e.g. trading rook for rook) moves should be ruled as a draw. If one side can legally capture the piece but the other side cannot legally capture the piece back, then it is considered a chase rather than an exchange. Pawns remaining on their side of the board that did not cross the river to the opponent's side cannot be chased.

\begin{algorithm}
\begin{algorithmic}[1]
\caption{get\_chases}\label{get_chases}
\Procedure{get\_chases}{$position$} 
\State $position.turn \gets \textbf{not}$ $position.turn$
\State $moves \gets$ \Call{gen\_captures}{$position$}
\State $chases \gets$ \Call{create\_move\_list}{}
\ForAll{$capture\_move \in moves$}
\If {\Call{pawn\_not\_passed}{$capture\_move$} \textbf{or} \textbf{not} \Call{is\_legal}{$position$, $capture\_move$}} 
\State \textbf{continue} \Comment{Skip moves which are not legal or involve a pawn crossing the river}
\EndIf
\If {\textbf{not} \Call{is\_exchange\_move}{$position$,$move$} \textbf{and} \textbf{not} \Call{is\_protected}{$position$, $move$}}
  \State \Call{add\_move\_to\_list}{$chases$, $capture\_move$}
\EndIf
\EndFor
\State $position.turn \gets \textbf{not}$ $position.turn$
\State \Return $chases$
\EndProcedure
\end{algorithmic}
\end{algorithm}

\subsection{Detecting a Chase}

If a check, capture move, or forward pawn move is detected, then we do not compute any chase information. Moves that result in a check are not chase moves since one is checking the king. Capture moves are irreversible moves that permanently lower the number of pieces in a given game. Because pawns can only move forwards without any ability to move backwards, pawn advances also cannot cause any repetitions.\\*

\begin{algorithm}
\begin{algorithmic}[1]
\caption{detect\_block}\label{detect_block}
\Procedure{detect\_block}{$position$,$move$,$chase\_moves$}
\If{\Call{in\_check}{$position$}}
\State \Return $Check$
\EndIf

\State $victims \gets \emptyset$ \Comment{Initialize victims as an empty set}

\ForAll{$chase\_move \in chase\_moves$}
  \State $to \gets chase\_move.to$
  \If{$chase\_move.to = move.from$}
    \State $chase\_move.to \gets move.to$
  \EndIf

  \If{\Call{is\_protected}{$position,chase\_move$}}
    \State $victims \gets victims \cup to$ \Comment{Add 'to' the set of victims}
  \EndIf
\EndFor

\If {$victims \neq \emptyset$}
  \State \Return $Chase$
\Else
  \State \Return $Idle$
\EndIf
\EndProcedure
\end{algorithmic}
\end{algorithm}

\noindent If a given position is in check, the position hash is labeled as a Check position. If a given position has been reached from a capture or pawn advance, its position hash is labeled a Cancel. Cancel is used to break from the loop early if an irreversible move is detected. If a position does not involve chasing, it is labeled Idle. If the position has a chase, it is labeled Chase.\\*

\begin{figure}[!tbp]
  \centering
    \includegraphics[width=0.3\textwidth]{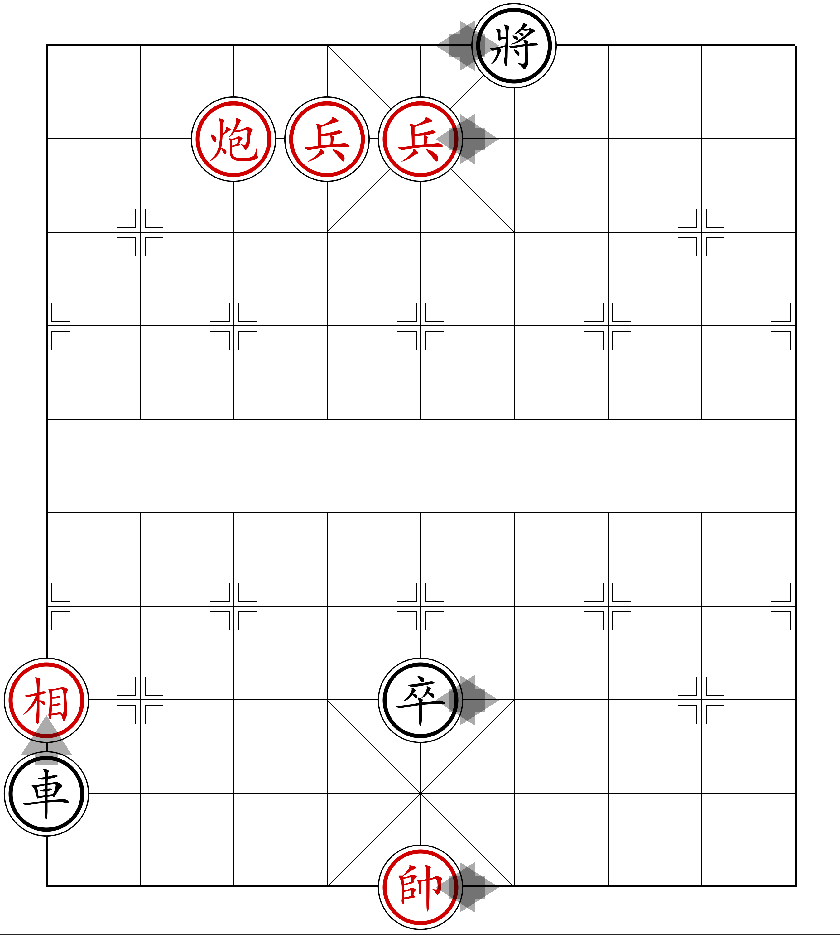}
    \caption{WXF Diagram 10: The black chariot can freely capture the undefended red elephant, but he does not since red is launching a mating attack. The black chariot is not considered chasing the red elephant since the red elephant is being sacrificed so red can launch a mating attack. Both sides are doing perpetual kill P5=4, K6=5, K5=4, P5=6, P4=5, K5=6...this will be ruled as a draw.}
\end{figure}

\noindent After generating captures on unprotected pieces, the algorithm determines whether the capture move is considered a chase. During a chase, the piece must be unprotected on one turn, and the next move responds to the threat by protecting the piece directly. If the opponent does not block or evade the chase and the piece can be taken for free, then it cannot be considered a chase, since the opponent did not respond to the chase (see WXF diagram 10, figure 8) [21]. An opponent may decide to counter-attack the chase by threatening a mating attack, sacrificing a piece, threatening to capture an opponent's piece, or accidentally giving up the piece for free. When these situations occur, the position should not be classified as a chase. Multiple pieces could be chased at the same time (highly unlikely in a normal game, but possible). If at least one piece is being chased, we mark the position as being chased.\\*

\noindent According to WXF rules [22], if multiple pieces are being chased by one or more pieces in a repetition, then it is ruled as a draw. If a single piece is being chased by one or more pieces in a repetition, it is considered a perpetual chase. It does not matter how many attackers are attacking a single piece. In this algorithm, we ignore the number of attackers and only focus on the set of chased victim pieces.\\*

\noindent To determine if a chase is occurring, we iterate through each possible capture move on an unprotected piece and check the opposing player's response to the capture move. If the opposing player's response to the capture is to evade the attack by moving, blocking, or protecting the piece, then we add the piece to a set of chased pieces. A chase is determined when a piece is unprotected on one turn and then protected on the next turn. If the opposing player's response does nothing to thwart the capture of the free piece, then it cannot be considered a chase.

\subsection{Detecting Repetition of a Player}

After labeling each position's hash and computing a set of pieces being chased, for every new position encountered that we want to check for a repetition, we iterate over the previous position hashes of the position hash history to check if there is a repetition for only a single player. If no repetition exists, this function returns undecided. If there is a repetition, this function attempts to classify the repetition violation level for a single player as either a perpetual check, perpetual chase, or perpetual idle. We have identified that all 16 different WXF repetition conditions fall into at least one of these 3 classes:\\*

\begin{algorithm}
\begin{algorithmic}[1]
\caption{judge\_player}\label{judge_player}

\algblock[Name]{Enum}{EndEnum}
\algblockdefx[NAME]{START}{END}%
[2][Unknown]{Start #1(#2)}%
{Ending}
\algblockdefx[NAME]{}{OTHEREND}%
[1]{Until (#1)}

\Enum{Violation\_Level}
\State Undecided = -1, PerpetualIdle = 0, PerpetualChase = 1, PerpetualCheck = 2
\EndEnum
\State

\Procedure{judge\_player}{$position\_history$,$index$,$ntimes$}
\If{count < 0 \textbf{or} {$position\_history[index].status = Cancel$}}
  \State \Return{Undecided}
\EndIf
\State $hash \gets position\_history[index].hash$
\State $index \gets index - 2$
\State $status \gets None$ \Comment{No positions have been analyzed yet. Set to None}
\State $repeating \gets False$
\State $initialized \gets False$
\State $chased\_set \gets \emptyset$ \Comment{Set the 'chased\_set' to the empty set, no chases detected so far}

\While{$index \geq 0$}
\If {$position\_history[index+1].status = Cancel$ \textbf{or} $position\_history[index].status = Cancel$}
  \State \textbf{break}
\EndIf
\State $status \gets status$ | $position\_history[index].status$ \Comment{'bitwise or' the position statuses together}
\If {$status = Chase$}
  \If{\textbf{not} $initialized$}
    \State $initialized = True$
    \State $chased\_set \gets position\_history[index].chased\_set$
  \Else
    \State \Call{update\_subset}{$chased\_set$, $position\_history[index+1].move$}
    \State $chased\_set \gets chased\_set \cap position\_history[index].chased\_set$
    \If{$chased\_set = \emptyset$}
      \State $status \gets Idle$
    \EndIf
  \EndIf
\EndIf

\If{$hash = position\_history[index].hash$} \Comment{Repetition Detected}
  \State $ntimes \gets ntimes - 1$
  \If {$ntimes = 0$} \Comment{If there are 'n' repetitions, set repeating to true}
     \State $repeating \gets True$
     \State \textbf{Break} 
  \EndIf
\EndIf

\EndWhile

\If{\textbf{not} $repeating$}
  \State \Return Undecided
\ElsIf{$Status = Check$}
  \State \Return PerpetualCheck
\ElsIf{$Status = Chase$}
  \State \Return PerpetualChase
\Else
  \State \Return Idle
\EndIf
\EndProcedure
\end{algorithmic}
\end{algorithm}

\begin{itemize}
\item \textbf{The perpetual check} is a repetition where the player performs a perpetual check only, without ever breaking off the check. Doing at least one check followed by at least one block or chase is not a perpetual check, but rather a perpetual idle.\\*

\item \textbf{The perpetual chase} is a repetition where a player performs a perpetual chase on a single unprotected piece other than a king. Any multiple attacking pieces can chase a single victim piece in this software implementation. If multiple victim pieces are being chased, it is not considered a chase, but rather a perpetual idle.\\*

\item \textbf{The perpetual idle} covers all other cases that are not perpetual checks or perpetual chases. Perpetual exchanges, offers, kills (threatening to checkmate the opponent), alternating between a check and kill, perpetually chasing two or more pieces, and alternating between a chase and check repetitions also fall into this category.\\*
\end{itemize}

\noindent As an overview of the whole algorithm, we start at the current hash of the current position, then traverse backward through the list of previous position hashes. We use bitwise or on the status flag enums to combine all the different statuses of positions together. If we hit a capture move, pawn forward move, or if there are no more previous positions, we break out of the loop. If the algorithm determines there is a perpetual chase, we initialize the chased set to the position victim set. Every time a piece is being chased, we update the piece position if the position has changed (e.g. a cannon evades an opponent chariot attack) and take the intersection between the current position chased set and the previous position chased set. If the same set of pieces is being perpetually chased consistently across all different positions, then the intersection of the chased pieces set is the same across different positions. If one set of pieces is being chased on one move and a completely different set of pieces is being chased on a different move, then both sets do not have anything in common and the intersection of sets will be empty. If the intersection of chased pieces is empty, label the repetition a perpetual idle, but if a set of pieces is being chased consistently across different positions, then label the repetition a perpetual chase.

\subsection{Judging A Chinese Chess Repetition}

In the main judge function, we take in as input a list of position hashes alongside a list of a set of chase victims for position hash. The judge function outputs a game result: undecided, win, loss, or draw relative to the player's turn. For example, if the function returns a win, it means the player whose turn it is won the game.\\*

\begin{algorithm}
\begin{algorithmic}[1]
\caption{judge\_ntimes}\label{judge_ntimes}
\Procedure{judge\_ntimes}{$position\_history$, $ntimes$}
\State $index \gets position\_history.count - 1$
\State $opponent\_violation\_level \gets$ \Call{judge\_player}{$position\_history$, $index$, $ntimes$}
\If {$opponent\_violation\_level = Undecided$}
\State \Return $Undecided$
\EndIf

\State $our\_violation\_level \gets$ \Call{judge\_player}{$position\_history$, $index-1$, $ntimes$}
\If {$our\_violation\_level = Undecided$}
\State \Return $Undecided$
\EndIf

\If {$opponent\_violation\_level = our\_violation\_level$}
  \State \Return $Draw$
\ElsIf{$our\_violation\_level > opponent\_violation\_level$}
  \State \Return $Loss$
\Else
  \State \Return $Win$
\EndIf

\EndProcedure
\end{algorithmic}
\end{algorithm}

\noindent The different outcome classes are represented in the code as enums. The ordering of these enum values from greatest to least are: perpetual check, perpetual chase, perpetual idle, and undecided. Perpetual check is the greatest violation, perpetual chase is in between, and perpetual idle is the least serious.\\*

\noindent In the main judge repetition function, we categorize the repetition of the first player, then categorize the repetition of the second player. If neither players repeat moves, then we return undecided. If the repetition of the player whose turn it is has a greater violation level than the violation level of the opposing player, then the player whose turn it is loses and we return loss. If the player whose turn is it has a violation level less than that of the opposing player, then it is considered a win. Otherwise, when the violation levels are equivalent, we return a draw.

\subsection{Testing The Repetition Rules}

We tested and verified that all repetition rules worked correctly by setting up positions found in the example diagrams, making the repetition sequences detailed in the example diagrams, and checking that the result of the main judge function matched the result described in the diagrams. We created 173 test cases using the 110 diagrams to test the win/draw/loss repetition sequences that could arise from each position.\\*

\noindent When integrating the repetition rules into the engine, we tested the engine to ensure that it handles repetitions properly by observing whether it handles the positions. In several of the WXF diagrams, there is a discrete checkmate solution and an optimal way to win the game based on the WXF repetition rules (see figure 4). In the diagrams with an optimal solution, we tested whether the engine could find the checkmate solution while handling repetitions correctly.

\section{Optimizations on the Repetition Rules in an Alpha Beta Negamax Engine}

The full implementation of chasing rules can be slow, and we utilize several techniques to speed up the repetition handling within the engine. These optimizations do not compromise the correctness of the engine.

\subsection{Staged Evaluation of Chasing Moves}

Alpha beta negamax chess engines rely heavily on good move ordering to prune the search tree aggressively and achieve higher depth. To briefly summarize the way a typical chess engine does move ordering, board moves are ordering with the principle variation best move first (there are many vast different kinds of implementations of this [11]), followed by capture moves, and last of all quiet moves. The vast majority of the time, a quiet move is refuted by a capture move since the majority of move sequences involve a piece moving into the attack range of an opponent's piece and losing that piece for free in the next turn.\\*

\noindent To illustrate the importance of giving capture moves higher priority in alpha beta negamax, one can compare the negamax performance debug logs of a didactic engine [20] without capture move ordering with one with a basic MVV-LVA (most valuable victim, least valuable attacker) capture move ordering heuristic.

\begin{verbatim}
info score cp 25 depth 1 seldepth 5 time 878 nodes 5161544
info score cp 25 depth 2 seldepth 5 time 6429 nodes 38369245
info score cp 20 depth 3 seldepth 5 time 121942 nodes 722743000
\end{verbatim}

\noindent Without MVV-LVA move ordering heuristic, the didactic Rustic Chess engine can only complete a depth of 3 over the course of 2 minutes. A lot of time is spent searching a negamax tree with a large branching factor.\\*

\begin{verbatim}
info score cp 25 depth 1 seldepth 3 time 0 nodes 1598
info score cp 25 depth 2 seldepth 3 time 1 nodes 3196
info score cp 20 depth 3 seldepth 5 time 2 nodes 7315
info score cp 20 depth 4 seldepth 5 time 6 nodes 20260
info score cp 5 depth 5 seldepth 9 time 25 nodes 76603
info score cp 20 depth 6 seldepth 7 time 109 nodes 293985
info score cp 5 depth 7 seldepth 9 time 543 nodes 1333835
info score cp 5 depth 8 seldepth 9 time 2900 nodes 7288058
info score cp -40 depth 9 seldepth 11 time 16933 nodes 39339223
\end{verbatim}

\noindent With MVV-LVA move ordering heuristic, the didactic Rustic Chess engine can complete a depth of 9 over the course of 2 minutes. This is a significant improvement from depth of 3 to a depth of 9, with many nodes being pruned quickly within the first 2-4 moves instead of searching through a negamax tree with a large branching factor. The engine can search much deeper and dramatically save millions of nodes just using MVV-LVA alone [20].\\*

\noindent Most engines use a staged move generation [17] that generates capture moves first, and delays generation of quiet moves until it is determined that the position is not refuted by a capture move beta cutoff. Because most nodes are refuted by the first few capture moves without ever searching quiet moves, the program can avoid generating all 35-45 different possible quiet moves if a beta cutoff occurs.\\*

\noindent Given that captures are usually given priority and cannot cause a repetition, we can delay computing chasing information of a position until the first quiet move is being evaluated. Since most nodes are pruned by a capture move on an unprotected piece, we can prune many negamax subtrees without computing chasing information. Since we compute chasing information during quiet moves, the program maintains its correctness. Algorithm 9 provides psuedo-code for how repetitions are evaluated within negamax.\\*

\begin{algorithm}
\begin{algorithmic}[1]
\caption{negamax}\label{negamax}
\Procedure{negamax}{$position$, $depth$, $\alpha$, $\beta$}
\State $result \gets$ \Call{judge}{$position$, $0$, $\beta$} \Comment{handle repetition}
\If {$result = Loss$} \State \Return $-20000$\Comment{Return Loss Score}
\ElsIf {$result = Draw$} \State \Return $0$ \Comment{Return Draw Score}
\ElsIf {$result = Win$} \State \Return $20000$ \Comment{Return Win Score}
\EndIf

\If {$depth = 0$} \Comment{base case}
  \State \Return \Call{evaluate}{$position$}
\EndIf

\State $chases \gets \emptyset$ \Comment{Only initialize chases if we are evaluating quiet moves}
\State $initialized \gets False$ \Comment{A boolean value tracking whether we initialized the chases}

\ForAll{$move \in position$}

  \If {\Call{is\_quiet}{$move$} \textbf{and} $initialized = False$}
    \State $initialized \gets True$
    \State $chases \gets$ \Call{gen\_captures}{}
  \EndIf
  \State \Call{make\_move}{$position$, $move$}
  \State $position\_history[ply].status \gets $\Call{detect\_block}{$positiion$, $move$, $chases$}

  \State $score \gets$ -\Call{negamax}{$position$, $depth-1$, $-\beta$, $-\alpha$}
  \State \Call{unmake\_move}{$position$, $move$}
  \If {$score \geq \beta$}
    \State \Return $score$
  \EndIf
  \If {$score > \alpha$}
    \State $\alpha \gets score$
  \EndIf
\EndFor

\State \Return $\alpha$
\EndProcedure
\end{algorithmic}
\end{algorithm}

\subsection{Judge Beta Cutoff}

In an alpha-beta negamax search, if an evaluation is greater than or equal to the beta value, then the search stops and prunes the tree. A good search depth is essential for being able to look deeply into a position and avoid a horizon effect, and getting an earlier beta cutoff is important for searching as deep as possible.\\*

\noindent In an alpha beta search, we are either looking for the optimal evaluation score for a position, or an evaluation score that is equal to or exceeds the beta cutoff value. If the draw score is equal to or exceeds the beta cutoff, just getting a draw score is good enough for accuracy. Algorithm 10 provides a pseudo-code description of how this works.\\*

\begin{algorithm}
\begin{algorithmic}[1]
\caption{judge\_prune}\label{judge_prune}
\Procedure{judge\_prune}{$position\_history$, $draw\_score$, $beta$}
\State $index \gets position\_history.count - 1$

\State $our\_violation\_level \gets$ \Call{judge\_player}{$position\_history$, $index-1$, $0$}
\If {$our\_violation\_level = Undecided$}
\State \Return $Undecided$
\EndIf

\If {$our\_violation\_level = Idle$ \textbf{and} $draw\_score \geq beta$}\Comment{If this case happens, take a faster shortcut}
  \If {\Call{is\_repetition}{$position\_history$, $index$}} \Comment{This shortcut cuts 5o\% of computation}
    \State \Return $Draw$
  \Else
    \State \Return $Undecided$
  \EndIf
\EndIf

\State $opponent\_violation\_level \gets$ \Call{judge\_player}{$position\_history$, $index$, $0$} \Comment{Regular Normal Path}
\If {$opponent\_violation\_level = Undecided$}
\State \Return $Undecided$
\EndIf

\If {$opponent\_violation\_level = our\_violation\_level$}
  \State \Return $Draw$
\ElsIf{$our\_violation\_level > opponent\_violation\_level$}
  \State \Return $Loss$
\Else
  \State \Return $Win$
\EndIf

\EndProcedure
\end{algorithmic}
\end{algorithm}

\noindent If a player is doing a perpetual idle in their turn, then we know that the player cannot lose, and the best that the player can get is either a win or a draw. If a draw is greater or equal to the beta cutoff, then we do not need to perform a check of the opponent's repetitions, but rather only a simpler check if the opponent is repeating moves. This saves us from a considerable amount of computation. In the condition described, we could skip 50\% of the computation.

\section{Results}

\subsection{Summary of Implementation Details of Engine}

\noindent Our prototype engine is a relatively powerful neural network negamax engine. Playing strength is expected to improve after training and learning from more games. The engine utilizes many complicated search techniques such as principle variation search [11], killer move ordering [1], capture move ordering, singular search extensions [2], null move pruning [7], countermove heuristics [19], late move reductions/pruning, static exchange evaluation [9], etc. to extend search in critical positions and prune nodes aggressively to get a high search depth. The evaluation function originally started as a basic handcrafted piece square table based off of [8], but eventually evolved into a NNUE engine trained on 50 million positions derived from 40000 grandmaster games from WXF Chinese Chess tournaments [21].

\subsection{Playing Style and Strength of Engine}

In terms of general playing style, integrating the complete implementation of Chinese Chess repetition rules into our engine caused the engine to attack opponents more aggressively and accurately rather than playing passively and defensively. Previously, the engine threw away critical attacking maneuvers because it was unsure about how to deal with repetitions. With the proper repetition rules implemented, it can understand how to take full advantage of the repetition rules to launch devastating attacks against opponents, in keeping with the aggressive attacking Chinese Chess strategies.\\*

\noindent We tested the engine by having it play against the previous version that only handled perpetual check without any perpetual chase handling. We tested the new engine using 300 games with a time control of 1 second with 0.20 second increment on a x86-64 6-core Intel Core i5-9600K with 3.70 GHz clock. Implementing repetition rules alone led to a gain of around +2 rating point increase, with an increase of 1\% winrate. After retraining the neural network based off of the evaluation scores affected by the complete WXF Chinese Chess repetition rules, we observed a +10 rating point increase, with a 5\% winrate increase. Throughout testing, no other changes were made to the engine's search, evaluation, or neural network training data. We used the Bayesian Elo [4] rating calculator program to measure the win/draw/loss ratio of the engine.\\*

\subsection{Playing Against Other Engines}

\noindent Elephant Eye [5] is a decently powerful open source Chinese Chess engine with an approximate rating of 2500. In September 2005, Elephant Eye participated in the 10th ICGA Computer Olympiad held in Taipei, with a record of 7 wins, 5 and 14 losses, ranking 11th out of 14 programs. In August 2006, Elephant Eye participated in the first national computer game championship held in Beijing, with a record of 7 wins, 2 losses and 11 losses, ranking 7th out of 18 programs. Elephant Eye has partial repetition coverage.\\*

\noindent When pitting our prototype engine against the Elephant Eye, we have observed several games where our prototype engine forced Elephant Eye to lose by repetition. Because Elephant Eye does not have full repetition coverage, our engine exploited Elephant Eye's weaknesses to obtain victory.\\*

\noindent Out of 100 games played between our engine and Elephant Eye played with a time control 1 second + 0.5 second increment on a x86-64 6-core Intel Core i5-9600K with 3.70 GHz clock, we won 38\% of the games, lost 41\% of the games, and drew 21\% of the games. The Bayesian Elo [4] rating calculator program estimates the rating of our engine to be about 9 rating points lower than Elephant Eye. With that being said, our prototype engine is decently powerful enough to stand up against relatively powerful engines such as Elephant Eye.

\section{Conclusion}
We fully implemented all WXF Chinese Chess repetition rules. It rules correctly for every possible example cases found within the WXF manual. Integrating this solution into our engine, the engine attacks more aggressively with a +10 point rating increase in playing strength. This work demonstrates that it is possible to fully integrate all WXF Chinese Chess repetition rules into an engine without dramatically slowing down the engine. The solution we present can be used to judge online games, serve as a backup referee, or be integrated into playing agents as we showcased in this paper.

\section{References}
\begin{enumerate}[label={[\arabic*]}]
\item Akl, S. G., \& Newborn, M. M. (1977, January). The principal continuation and the killer heuristic. In Proceedings of the 1977 annual conference (pp. 466-473).

\item Anantharaman, T., Campbell, M. S., \& Hsu, F. H. (1990). Singular extensions: Adding selectivity to brute-force searching. Artificial Intelligence, 43(1), 99-109.

\item China Xiangqi Association. The Playing Rules of Chinese Chess (in Chinese). Shanghai Lexicon Publishing Company (1999) IBSN: 7-5326-0556-6

\item Coulom, R. (2010, March 31). Bayesian Elo Rating. Bayesian Elo rating. https://www.remi-coulom.fr/Bayesian-Elo/

\item Elephant Eye. (2008, March). https://github.com/xqbase/eleeye

\item Fichter, F. (2021, November 19). Fairy-Stockfish. https://fairy-stockfish.github.io/

\item Goetsch, G., \& Campbell, M. S. (1990). Experiments with the null-move heuristic. In Computers, Chess, and Cognition (pp. 159-168). New York, NY: Springer New York.

\item Li, C. (2008). Using AdaBoost to Implement Chinese Chess Evaluation Functions (Doctoral dissertation, University of California, Los Angeles).

\item Michie D (1966): Game Playing and Game Learning Automata. In Advances in Programming and Non-Numerical Computation, Ed. Fox L, pp 183-200. Oxford, Pergamon.

\item Nasu, Y. (2018). Efficiently updatable neural-network-based evaluation functions for computer shogi. The 28th World Computer Shogi Championship Appeal Document, 185.

\item Pearl, J. (1980, August). SCOUT: A Simple Game-Searching Algorithm with Proven Optimal Properties. In AAAI (pp. 143-145).

\item Pham, N. H. (2018). A completed implementation for Xiangqi rules. ICGA Journal, 40(3), 305-317.

\item Pikafish. (2022, June 28). https://pikafish.org/

\item Png, J. (n.d.). The Rules of Xiangqi (Chinese Chess). xqinenglish.\\*https://www.xqinenglish.com/index.php?option=com\_content\&view=article\&id=923\%3Athe-rules-of-xiangqi-chinese-chess\&catid=119\&Itemid=569\&lang=en

\item Png, J. (n.d.). xqinenglish - Home. www.xqinenglish.com. Retrieved February 7, 2024, \\*from https://www.xqinenglish.com/index.php?lang=en

\item Silver, D., Hubert, T., Schrittwieser, J., Antonoglou, I., Lai, M., Guez, A., ... \& Hassabis, D. (2018). A general reinforcement learning algorithm that masters chess, shogi, and Go through self-play. Science, 362(6419), 1140-1144.

\item Stockfish Open source chess engine. Stockfish. (n.d.). https://stockfishchess.org/ 

\item Tang, P., Castillo, E., \& Pai, J. T. (1998). Xie Xie Master. XieXie. http://xiexiemaster.com/home.php 

\item Uiterwijk, J. W. (1992). The countermove heuristic. ICGA Journal, 15(1), 8-15.

\item Vanthoor, M. (2021, January 24). MVV\_LVA. Creating the rustic chess engine. https://rustic-chess.org/search/ordering/mvv\_lva.html 

\item World Xiangqi Federation. (n.d.). Past WXF tournaments game records. World Xiangqi Federation.\\*https://www.wxf-xiangqi.org/index.php?option=com\_content\&view=article\&id=218\&Itemid=313\&lang=en

\item World Xiangqi Federation. (2018, April). World Xiangqi Rules. wxf-xiangqi.org. https://www.wxf-xiangqi.org/images/wxf-rules/2018\_World\_XiangQi\_Rules\_English2018.pdf

\item Wu, K. C., Hsu, S. C., \& Hsu, T. S. (2006). The graph-history interaction problem in chinese chess. In Advances in Computer Games: 11th International Conference, ACG 2005, Taipei, Taiwan, September 6-9, 2005. Revised Papers 11 (pp. 165-179). Springer Berlin Heidelberg.

\item Xu, S.-Y.: Xiangqi Qili Yu Daipan Jumian De Caijue (Rulings of Chinese Chess Games that are not Clearly Stated in the Current Rules). People’s Athelete Pulishing Co. (2000) (in Chinese) ISBN: 7-5009-1925-5

\item Xqbase. (2015, February 6). Xiangqi Wizard Light. GitHub. https://github.com/xqbase/xqwlight

\item Zobrist, A. L. (1990). A new hashing method with application for game playing. ICGA Journal, 13(2), 69-73.

\end{enumerate}

\end{document}